\documentclass[final,5p,twocolumn]{elsarticle}

\usepackage[utf8]{inputenc}
\usepackage{amsmath, amssymb, amsfonts, bm, amsthm}
\usepackage{algorithm}
\usepackage{algpseudocode}
\usepackage{graphicx}
\usepackage{xcolor}
\usepackage{booktabs}
\usepackage{url}
\usepackage{hyperref}

\journal{.}

\begin{document}

\begin{frontmatter}

\title{Deep learning framework for crater detection and identification on the Moon and Mars}

\author[2]{Yihan Ma}
  
\author[2]{Zeyang Yu}

\author[2]{Rohitash Chandra \corref{cor1}}
\cortext[cor1]{Email: rohitash.chandra@unsw.edu.au}  
 
 
\affiliation[2]{Transitional Artificial Intelligence Research Group, School of Mathematics and Statistics, UNSW Sydney, Sydney, Australia}


\begin{abstract}
Impact craters are among the most prominent geomorphological features on planetary surfaces and are of substantial significance in planetary science research. Their spatial distribution and morphological characteristics provide critical information on planetary surface composition, geological history, and impact processes. 
In recent years, the rapid advancement of deep learning models has fostered significant interest in automated crater detection. In this paper, we apply advancements in deep learning models for impact crater detection and identification. We use novel models, including Convolutional Neural Networks (CNNs) and variants such as YOLO and ResNet. We present a framework that features a two-stage approach where the first stage features crater identification using simple classic CNN, ResNet-50 and YOLO. In the second stage, our framework employs YOLO-based detection for crater localisation. Therefore, we detect and identify different types of craters and present a summary report with remote sensing data for a selected region. We consider selected regions for craters and identification from Mars and the Moon based on remote sensing data. Our results indicate that YOLO demonstrates the most balanced crater detection performance, while ResNet-50 excels in identifying large craters with high precision.   

\end{abstract}


\end{frontmatter}

\section{Introduction}
 
The automatic detection of craters is a fundamental task in planetary science and has significant implications for geological analysis \cite{Shirmard2022}, spacecraft navigation \cite{Erdem2022}, and planetary surface exploration \cite{Yang2020}. The identification of craters is essential for spacecraft navigation, identifying hazardous terrains, and exploring planetary resources. Traditional crater identification methodologies primarily rely on manual visual interpretation, which is labour-intensive, time-consuming, and susceptible to subjective biases.  Crater detection methods can be broadly categorised into traditional computer vision and machine learning  \cite{Di2014} and deep learning-based approaches \cite{Xiong2025}. Traditional methods typically rely on manually designed features such as edges, contours, and shaded regions, employing template matching and machine learning classifiers for crater identification. However, these methods face substantial limitations when dealing with craters of varying sizes and morphological characteristics. In contrast, deep learning methods eliminate the need for manual feature extraction by learning data representations automatically through end-to-end training\cite{Lecun2015DL}. This significantly enhances the accuracy and adaptability of crater detection\cite{tewari2023deeplearning}.  For example, Silburt et al. \cite{Silburt2018} applied U-Net to lunar data sets and achieved high levels of precision in their attempts to detect craters. Yang et al. \cite{Yang2021HRFPNet} used high-resolution feature pyramid networks for the detection of small-scale craters, which enabled the crater detection task to be effective for small targets.

Conventional computer vision methods relied on manual feature extraction techniques (e.g., edge detection and template matching). Deep learning has revolutionised computer vision, enabling significant advances in tasks such as image classification, object detection, and segmentation \cite{Lecun2015DL}. Deep learning models are capable of autonomously learning and extracting relevant features directly from image data \cite{li2018survey}. Deep learning models have demonstrated superior accuracy and generalisation capability when applied to large-scale, high-resolution remote sensing data sets \cite{li2019hyperspectral}. 
Deep learning models such as Convolutional Neural Networks (CNNs)\cite{LeCun1998cnn} are widely used in visual recognition tasks such as image classification, face recognition, image denoising and scene recognition. Hence, CNNs with semantic segmentation models have been used for high-resolution identification and characterisation of craters from imagery \cite{Silburt2018}. Furthermore, CNN variants such as  U-Net\cite{Ronneberger2015UNet} and Faster R-CNN\cite{ren2015faster} have been fine-tuned to work well even with the constant change in sizes, shapes, and degradation states of the craters \cite{Cui2020DegradationReview}. Out of which, CNN-based models such as  YOLO (You Only Look Once)\cite{Redmon2016YOLO} have been used for real-time object detection with high accuracy.

Residual Networks (ResNet) \cite{he2016deep} have been prominent for classification tasks and are often used as the backbone feature extractor for YOLO.
ResNet and its variations (such as ResNet-50) are an enhanced deep CNN featuring the addition of residual connections that eliminate the common problem of gradient vanishing in conventional CNNs. This can achieve a deeper network structure and greatly increase feature expression capability and accuracy in recognition. It is quite common for researchers to employ ResNet-50 in image classification and feature extraction networks for target detection, image segmentation due to its deep residual architecture, which enables effective training of deep networks while mitigating vanishing gradient issues \cite{ramasamy2021multiclass}. It also performs quite well in more complicated visual tasks, such as medical diagnoses \cite{chen2021medical} and industrial applications \cite{sahoo2022manufacturing}. Since the crater shape varies and has fuzzy boundaries, traditional CNNs have difficulties in extracting complex features effectively. There are several studies where deep learning and machine learning models have been trained using data from NASA (National Aeronautics and Space Administration) for crater detection \cite{tewari2023crater, delprete2022crater}.
ResNet has the potential to do better in deep background interference due to its deep residual structure, hence a trend to improve robustness and accuracy of classification. Fast R-CNN \cite{ren2015faster} serves as a two-stage target detection model, and provides an improvement by sharing the convolution feature and regressing candidate boxes to improve the accuracy and efficiency of target detection. It significantly covers moderately realistic scenarios with high precision, such as the applications of remote sensing-based target location \cite{zhao2022resnet}, automatic driving traffic sign recognition \cite{li2023traffic}, and medical focus detection tasks \cite{zhao2020dares2net}.

Furthermore, machine learning paradigms such as transfer learning \cite{Pan2010TransferLearning} enabled models trained for low-resolution datasets to be quite effective in high-resolution datasets. Yang et al. \cite{Yang2020} utilised transfer learning for the study of the moon with the data from the Chinese Chang’E missions. Using Chang’E-1 and Chang’E-2 data \cite{Silburt2018}, Yang et al. \cite{Yang2020} applied deep and transfer learning to identify over 117,000 lunar craters and estimated ages for nearly 19,000. The authors utilised a two-stage CNN-based model for multiscale detection and age classification by integrating morphological and stratigraphic features with high accuracy. Undoubtedly, these models would not just continue to expand and sharpen the catalogues of craters, but also provide a background on the impact history and geological development necessary to optimally conduct space exploration and mission planning.

A major difficulty in crater detection is the problem of data labelling because quality labelled datasets are relatively rare; thus, the good old manual annotation is laborious and costly. The other problem is scale variation, since the diameter of craters is in the range of a few meters to a few hundred kilometres, it becomes imperative that the models should be able to handle very different scales effectively. Finally, topographic complexity has proven to be a stumbling block toward generalisation in the context of the different surface features of different planetary bodies \cite{di2014crater}. Hence, models trained on one dataset do not perform well on others, especially across different planetary bodies, due to differences in data resolution, lighting conditions, and terrain features \cite{tewari2023deeplearning} 
Limited computational resources and class imbalance also hinder model performance, especially when training deep networks or detecting rare crater types in unbalanced datasets. However, novel deep learning models such as ResNet and YOLO have the potential to address some of these limitations. 

In this paper, we present a robust deep learning framework for the automation of crater detection and analysis. Our framework enables us to evaluate novel deep learning models (ResNet, CNN, and YOLO) for crater detection and identification, which is a two-stage process, whereby YOLO is used for detection, and refined CNN models are used for recognition. We provide an annotated crater detection dataset, which is utilized for training the respective models for crater recognition, where large, small, and medium craters are categorized.  The framework intends to enable us to improve our knowledge of the planetary surface processes, which are two of the fundamental aspects recommended in critical tasks of space exploration, such as landing site selection and resource assessment. Finally, our framework enables us to generate a report that provides statistics about the types of craters in a given region. We consider selected regions for craters and identification from Mars and the Moon based on remote sensing data.

The rest of this paper is organised as follows. Section 2 provides a background on crater exploration, and Section 3 introduces the methodology and includes data processing and modeling. In Section 4, we present the results, followed by a discussion in Section 5. Finally, we conclude the paper in Section 6.

\section{Background}

\subsection{Deep learning in planetary crater identification}

In planetary science, deep learning models have been increasingly adopted for tasks such as crater detection, terrain classification, and rock identification \cite{Silburt2019}. Traditional methods for crater detection, which often rely on hand-crafted feature extraction techniques such as edge detection and template matching, struggle with challenges including noise, varying illumination, and the diverse morphology of craters \cite{urbach2009automatic}. In contrast, deep learning models, particularly  CNNs have demonstrated the ability to automatically learn relevant features from raw data, making them highly effective for crater identification tasks \cite{He2015DL}.  CNNs can automatically learn to detect features such as crater rims, shadows, and textures, which are critical for distinguishing craters from other geological formations \cite{krizhevsky2012alexnet}. The ability of CNNs to generalise from large datasets has made them a popular choice for planetary science applications, where manual feature extraction is often impractical \cite{wu2021monitoring}. 

For example, Silburt et al. \cite{Silburt2018} employed CNNs to classify lunar craters based on their morphological features, achieving robust performance even in complex terrains. Similarly, the YOLO model has been utilised for real-time crater detection in planetary exploration missions, offering a balance between speed and accuracy \cite{Redmon2018YOLOV3}. These advancements highlight the potential of deep learning to revolutionise planetary science by automating and improving the accuracy of crater detection and analysis. However, CNNs require large amounts of labelled data for training, and their performance may degrade when applied to datasets with significant domain shifts or limited annotations





Unlike traditional two-stage detectors, the YOLO model \cite{Redmon2016YOLO} predicts bounding boxes and class probabilities directly from the input image in a single forward pass, enabling real-time performance \cite{redmon2016yolo9000}. This efficiency makes YOLO particularly suitable for applications requiring fast processing, such as onboard spacecraft systems or large-scale planetary image analysis \cite{howard2017}. However, the single-stage design of  YOLO can sometimes result in lower precision for small or densely packed objects, which may require additional optimisation for specific tasks such as  crater detection \cite{lin2017focal}. 


ResNet-50 is a 50-layer implementation of the ResNet (Residual Network) architecture, which skip connections to help mitigate the vanishing gradient problem in deep networks \cite{He2015DL}. These connections allow the network to learn residual mappings, enabling the training of very deep architectures without degradation in performance \cite{he2016identity}. ResNet-50 has demonstrated exceptional results in image classification tasks \cite{he2016deep}, particularly in scenarios requiring the recognition of complex patterns, such as crater identification \cite{huang2016densely}. Using pre-trained weights from large datasets such as ImageNet, ResNet-50 can achieve high accuracy even with limited planetary data, making it a powerful tool for crater detection and classification \cite{yosinski2014transferable}. However, the computational cost of ResNet-50 can be high, and its performance could be limited when applied to very high-resolution images or datasets with significant class imbalance.

\section{Methodology}
 
\subsection{Data}
\subsubsection{NASA}

NASA  is a  United States federal agency popularly known for managing the space programs and spearheading aerospace and space science research. NASA is well known for sharing data with national and international research institutions and has been at the forefront of space research globally.

The dataset utilised in this study consists of high-resolution satellite images of planetary surfaces, primarily focusing on Mars and Mercury. These images were sourced from NASA's publicly accessible archives, which include data from missions such as the Mars Reconnaissance Orbiter (MRO) \cite{McEwen2007}, and the MESSENGER (MErcury Surface, Space ENvironment, GEochemistry, and Ranging) spacecraft\cite{Hamelin2007}. These missions have provided extensive visual data of planetary surfaces, capturing a wide range of geological features. The dataset encompasses images with diverse resolutions, varying lighting conditions, and different surface characteristics, making it well-suited for training and evaluating deep learning models aimed at crater detection and classification tasks \cite{Silburt2018}.

\subsubsection{Roboflow Universe}

Roboflow Universe \footnote{\url{https://universe.roboflow.com/}} is a community-driven collaborative platform for the computer vision community to provide high-quality datasets \cite{ciaglia2022roboflow} and pre-trained models. Roboflow Universe assembles thousands of openly sourced, annotated, and collated datasets to enable users to circumvent the mundane process of data preparation and cleaning. Object detection, image segmentation, classification, and object tracking are examples of tasks supported.  It further aggregates most of the widely used models,  such as YOLO and Faster R-CNN. The pre-trained models can be downloaded by users or used to experience online for the model effects to enable fast prototyping and application deployment.  The Roboflow Universe also has a full API interface and a simple Source Development Kit (SDK) to support quick integration and deployment of models for web, mobile, or edge devices, truly achieving one-stop computer vision development.   Therefore, in this study, Roboflow will help in building a knowledge-sharing and project-sharing community among peers because this can be freely accessed by the data sets of lunar crater images formed or shared by others. The data of lunar craters on Roboflow Universe have been taken from different moon exploration missions and varied shooting conditions.
The diversity of data richness can effectively improve the generalisation performance of the training model and enhance the robust nature of the model in the actual scene.


\subsection{Manual data labelling}


We analyse crater datasets from Mars and the Moon, which were obtained from NASA PDS\cite{NASApds_mars}. The Mars region covers the mid-low southern latitudes (Latitude: -25.40° to -25.17°, Longitude: 196.23°E to 196.34°E), while the Moon region covers the western edge of the equatorial region on the near side (Latitude: -1.60° to -1.27°, Longitude: 318.04°E to 318.16°E). We categorized the craters into three size groups based on their diameter: small (\textless10 km), medium (10-50 km), and large (\textgreater50 km). The Mars dataset contains 135 small, 20 medium, and 16 large craters (171 total). The Moon dataset includes 611 small, 45 medium, and 15 large craters (671 total). We provide our labelled data using a GitHub repository \footnote{\url{https://github.com/sydney-machine-learning/crater-identification}}. 







\begin{figure*}[h]
     \centering
    \includegraphics[width=\linewidth]{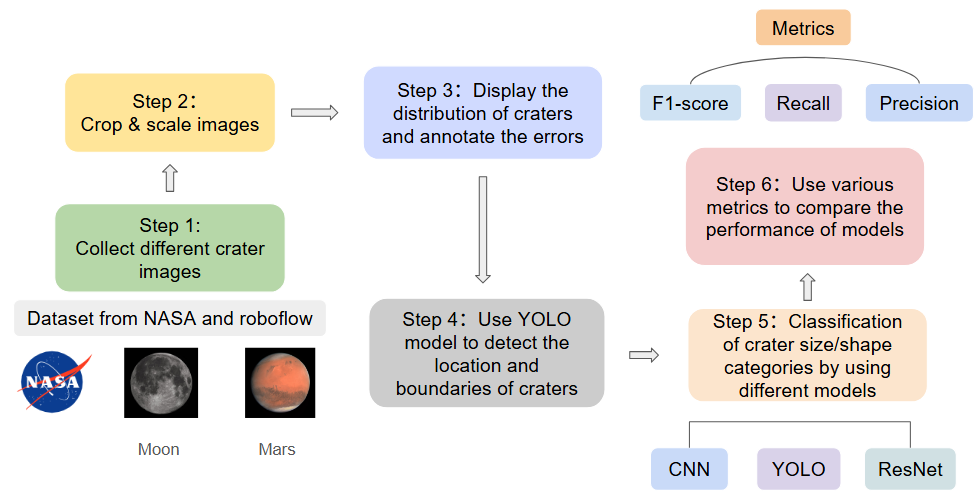}
    \caption{Framework of crater detection and identification, which features a two-stage process that requires two separate deep landing models.}
    \label{fig:Framework}

\end{figure*}

\subsection{Framework}

We consider selected regions for craters and identification from Mars and the Moon. 

Our framework features a more advanced and intricate YOLO (namely YOLO-v11) for crater detection. We use two deep learning models for large remote sensing image datasets.



We highlight that conventional deep learning models, such as CNN for object detection and identification, take a fixed window size of the input image. This can be problematic in real applications such as crater detection, where the object (crater)  size varies and does not fit in a single input image. In such cases, an input image may feature incomplete craters, which would be difficult to identify. Moreover, we have a two-stage problem, where in the first stage, our framework needs to detect the crater and in the second stage, it identifies the type of crater, eg. small, medium and large craters, or incomplete craters. 

 We refer to relevent study to build an effective crater detection framework. Neubeck and Van Gool \cite{neubeck2006efficient}, presented the Non-Maximum Suppression (NMS) method, which helps us remove overlapping boxes and improve detection accuracy. In this way, we handle the problem of different crater sizes and crowded regions in planetary images more effectively.

\autoref{fig:Framework} presents the framework with three deep learning models: CNN, YOLO, and ResNet-50 for detecting and identifying craters on Mars and the Moon. 

The first step was about data acquisition and preprocessing. Our dataset for this study came from two different places. First, we got high-resolution images of the Martian surface from the NASA official website \cite{nasa_data}. Second, images related to lunar craters were from the Roboflow public platform \cite {roboflow2024}. Since the image sizes were quite different, we preprocessed each one specifically. The Mars images were cropped, resized, and then relabelled to adapt to the model's input requirements. Lunar images were padded with black to be enlarged and then relabelled to have a uniform image size.

In the second step, we perform data visualisation and annotation correction. To assure data quality and validity, we will carry on detailed data visualisation analysis on all processed data to observe the distribution characteristics of craters. Also, we will pay special attention to the bias problems manuall annotation may have, such as large craters being incorrectly labelled as medium-sized ones. We then correct and relabel them one by one. At this point, we paid special attention to the bias problems that manual annotations had caused, such as large craters being incorrectly labelled as medium-sized craters. The accuracy and stability of the dataset have to be maintained in due consideration in this respect, or consideration of this bias.

Step three involved conducting YOLO model training and prediction. We chose the YOLO model for model training with two kinds of classification experiments: three-class classification (large, medium, and small craters) and two-class classification (craters versus background). In order to facilitate the ability of YOLO to process large images, two different prediction strategies were designed and implemented:

\begin{enumerate}
  
 \item  Direct prediction method: The ultra-clear 4K images were input into the YOLO model as a whole for detection. This approach has obvious advantages since it is able to carry out efficient and accurate detection of large-sized craters without any need for additional overlapping detection boxes. However, its accuracy in identifying small-sized craters is not that high.

 \item  Sliding window prediction method: An image was broken down into many small regions of 640x640 for finding individual detection with a 30 percent overlap of regions so that no targets are missed. This helped increase the recognition accuracy of small-sized craters a lot, but it has obvious deficiencies in large-sized crater detection. We further adopted the Non-Maximum Suppression (NMS)\cite{neubeck2006efficient} technique, which removes redundant detected boxes created by the sliding window method, therefore, efficiency and accuracy in the end-to-end prediction.
 
\end{enumerate}

In the fourth step, we conduct benchmark experiments with the CNN model. In further model comparison, the traditional CNN model was used as a benchmark for crater classification experiments. A dedicated data loading and preprocessing process was developed which prepared data for the CNN model and it was in the YOLO dataset format non-crater images were separately labeled. The CNN model was developed in the Keras framework with an architecture that included three convolutional layers, pooling layers, and a fully connected layer with 128 neurons, and a Softmax output layer\cite{goodfellow2016deep}. Model training is done by us where we introduced data augmentation strategy that includes random rotation and horizontal flipping to add the generalization ability of the model. Setting model checkpoints and early stopping mechanisms also help us prevent the overfitting of the model.

In the fifth step, we conducted the experiment and made a discussion for future work. The model was evaluated using F1 score, precision, recall, and accuracy metrics. Thirty independent experiments were conducted to calculate the mean and variance of the results to ensure statistical significance. Therefore, the present study objectively indicates that the CNN model is shallowly optimized and used as a benchmark for performance evaluation. These are the reasons for this research, as described henceforth in an objective comparison of the actual detection performance of advanced models, such as YOLO and ResNet50, with the benchmark performance of the CNN model. 

In the future, we will continue with the in-depth research of more complex network architectures, like ResNet50, and take into account advanced fusions such as model fusion and transfer learning\cite{tan2018survey} to further advance the accuracy and universality on crater detection, subsequently leading to the refined development of space exploration and geological research.

\subsection{CNN Model}
\begin{figure*}[htbp!]
    \centering    \includegraphics[width=1\linewidth]{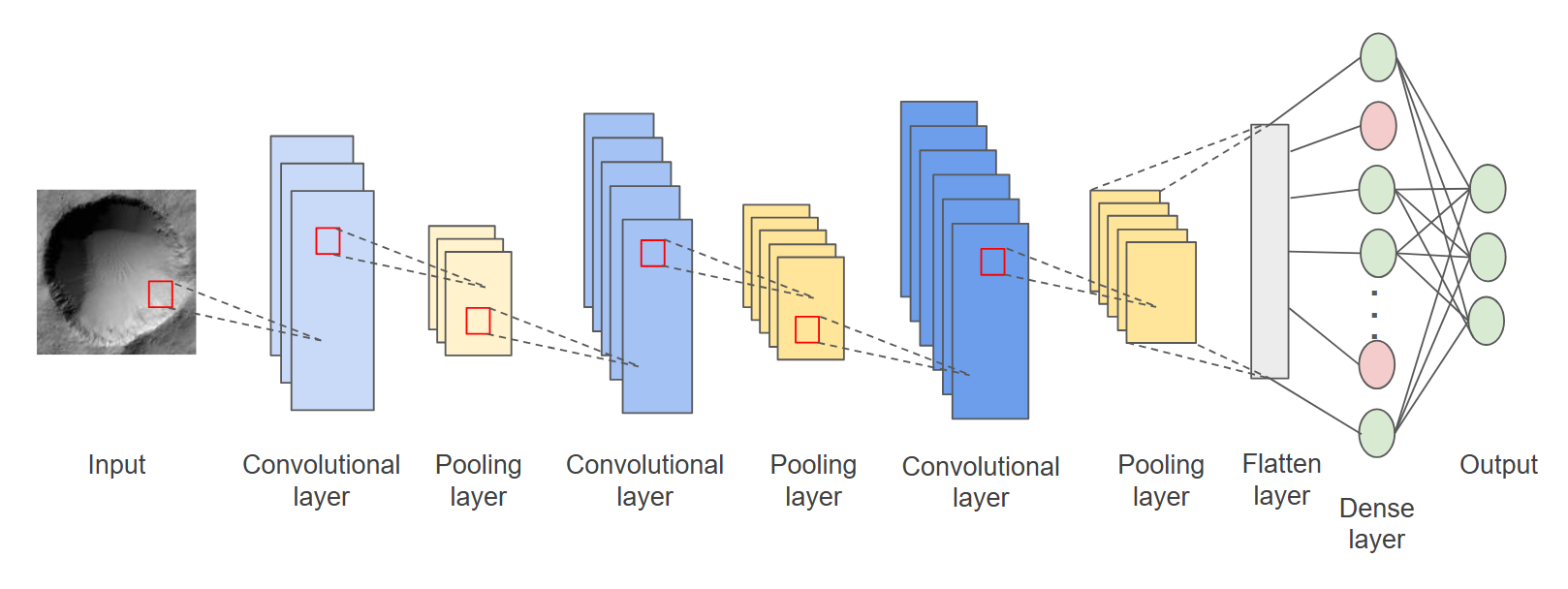}
    \caption{CNN Architecture}
    \label{fig:CNN}    
\end{figure*}

\autoref{fig:CNN} is our process using the CNN model as a benchmark model for the classification task of crater images. To compare the performance of more advanced models such as YOLO and ResNet-50\cite{he2016deep}, based on a data set in YOLO format, we crop the crater region in the original image into a single image of 128×128 pixels. We also introduce the "non-crater" category to construct a four-classification task. The data is organized in structured training, validation, and testing directories, and loaded and enhanced by PyTorch Dataset class\cite{paszke2019pytorch} and Keras ImageDataGenerator\cite{chollet2015keras}. This includes pixel value scaling random rotation and horizontal flip operations to enhance the robustness and generalisation ability of the model. The data is then formatted for the model input and output, pixel values of the images and their labels converted into tensor forms for faster computation, also scaled by a data scaler, and finally transformed into a tuple containing both the image and its label for the convenience of the users of the model. The input shape for the model is determined to be 128x128x3 by the size of the images, which are standardised to be 128 pixels in height and 128 pixels in width with three colour channels. The model was trained with Adam optimiser for 30 rounds and categorical crossentropy used as loss function\cite{kingma2014adam}. The final results were evaluated by accuracy, loss, classification reports, and confusion matrix.

\subsection{YOLO Model}
\begin{figure*}[htbp!]
   \centering
   \includegraphics[width=12cm]{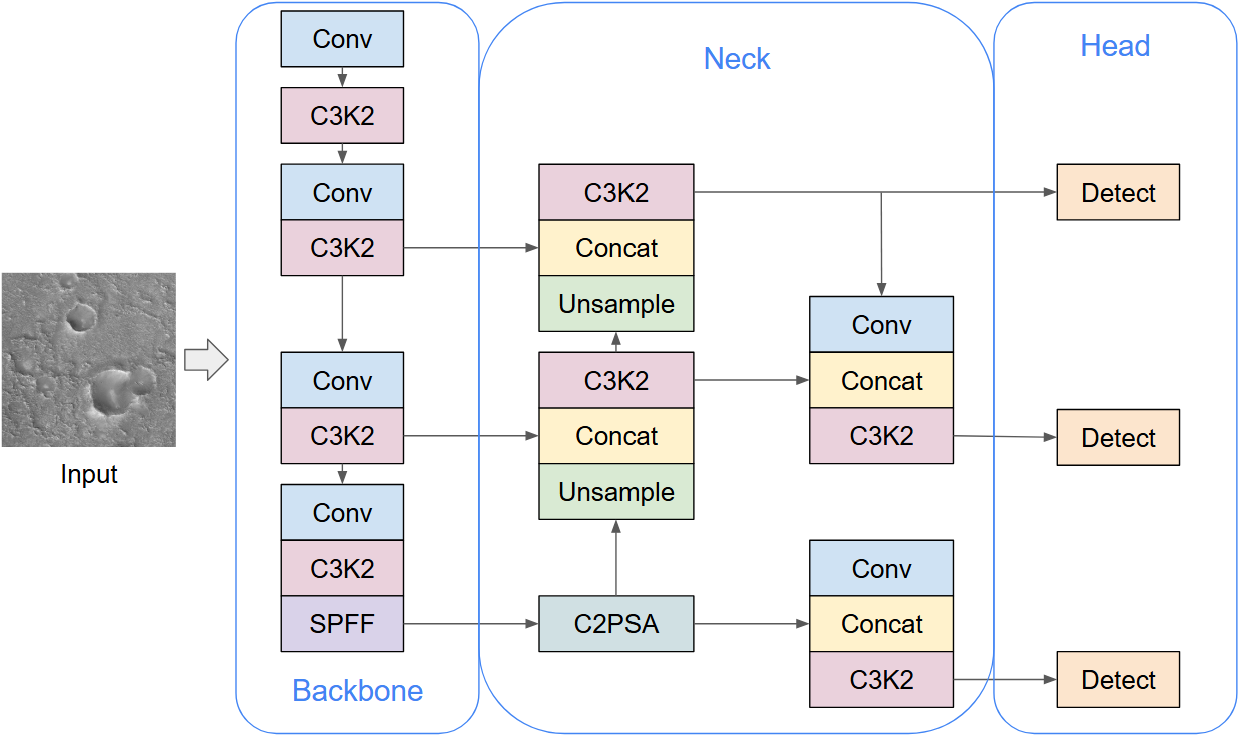}
   \caption{YOLO Architecture}
    \label{fig:YOLO}
\end{figure*}

\autoref{fig:YOLO}is our process using the YOLO method.We all have different priorities and ways of using YOLO. In the first focus, YOLO primarily serves as a data annotation tool since the original dataset is first annotated in YOLO format, because more traditional CNN models cannot directly process YOLO data with a background (non-crater) category. An image marked by YOLO format is then cropped to show a single crater and resized to 128x128 pixels suitable for CNN training. Here, YOLO will mainly undertake accurate positioning information and data preprocessing; The second focus goes directly into the application of the YOLO model itself, wherein not only the use of the latest version of YOLO (YOLOv11) for training and classification detection of crater image data but also for large, medium, and small craters of different sizes. Also, two strategies for predicting ultra-large 4K resolution planetary surface images are offered; one where the complete large image is input to YOLO for fast and efficient large impact crater detection, and the other where the large image is gradually segmented into several 640x640 small regions using sliding window\cite{zhao2020sliding}. After making individual predictions for each region, the following two strategies will be proposed: the Non-Maximum Suppression (NMS) methodology for consolidating overlapped detection boxes\cite{neubeck2006efficient}, hence raising small crater detection precision and accuracy by a large measure. Thus, in general, the YOLO in the first focus mainly acts as data pretreatment and gives CNN training data, while the YOLO in the second focus serves as the principal model for extensive multi-scale crater detection applications. And via the well-designed foretell strategy and post-processing means, the advantages and strong applicability of YOLO in real deep learning applications are demonstrated.

\subsection{ResNet-based Model}
\begin{figure}[H]
   \centering
   \includegraphics[width=\linewidth]{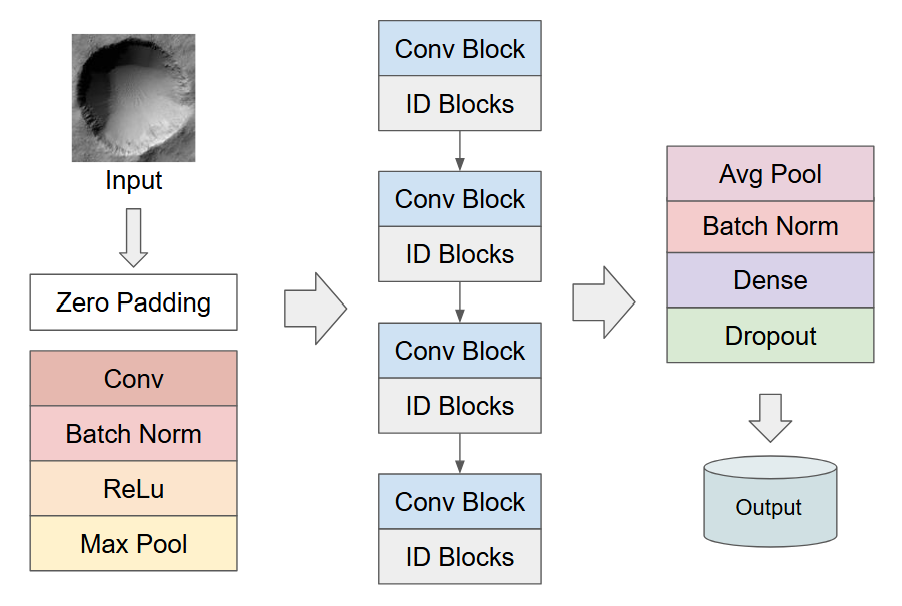}
   \caption{ResNet Architecture}
    \label{fig:ResNet50}
\end{figure}

The   impact crater identification is articulated through the ResNet-50  deep learning model as shown in  \autoref{fig:ResNet50} \cite{he2016deep}. An image is first preprocessed using YOLO marker coordinates, then the area of the crater is cropped and fixed at 128x128 pixels. Next, real-time data augmentation is performed by Keras  \footnote{ImageDataGenerator: \url{}} to improve the generalisation of the model. The model itself is based on a pre-trained ResNet50, only the top layer structure is customised. Global average pooling and batch normalisation, a full connection layer with L2 regularisation, and a Dropout layer are used to reduce the overfitting risk. The real business data is from OUC and has about four types of targets to classify by the Softmax in the output layer. We ignore the non-craters for analysis. In the training process, Adam optimiser and classification cross entropy loss function\cite{kingma2014adam} were used to monitor the performance of the model in real-time by custom F1-score, and an adaptive learning rate strategy is applied to get the best performance. The final results show the accuracy and stability of the model based on 30 experimental statistical evaluations.

\subsection{Technical details}
Our crater discovery system unites three models: YOLO-v11\cite{ultralytics2023} for spotting objects and ResNet50 plus a usual CNN for labeling craters. In the case of YOLOv11, we take a project-included model and begin training from zero. While ResNe50 comes shot up with ImageNet pre-trained weights and during training\cite{he2016deep}, all layers of convolutions are stopped. That is, our own CNN model is built by hand and trained from zero without any pre-trained weights. These models are used to cross and compare the performance between the discovery and labeling tasks.

The work was carried out using Python, employing TensorFlow Keras\cite{tensorflow2015-whitepaper} for the models in CNN and ResNet50, and the Ultralytics library for YOLOv11, with supporting help from NumPy\cite{harris2020array}, OpenCV\cite{bradski2000opencv}, Matplotlib\cite{hunter2007matplotlib}, Seaborn, and Plotly. Training has been performed on a workstation that possesses an NVIDIA GeForce RTX 4060 GPU and 16 GB memory.

A custom dataset named datasets was created, comprising three crater categories: large, medium, and small. These data come from Mars and the Moon, respectively. We created two versions of the dataset as required by YOLO  for object detection and in cropped format (based on YOLO bounding boxes) for classification models. It was further divided into training (60\%), validation (10\%), and test (30\%) sets with the following number of samples in each for the case of
Mars craters as shown in Table \ref{tab:crater_distribution}. Table \ref{tab:crater_distribution-moon} presents the distribution of the craters for the case of the Moon.

  

\begin{table}[htbp!]
\centering
\caption{Distribution of craters by size in dataset of Mars}
\begin{tabular}{|c|c|c|c|}
\hline
\textbf{Dataset} & \textbf{Large} & \textbf{Medium} & \textbf{Small} \\
\hline
Training Set     & 54   & 68  & 508\\ \hline
Validation Set   & 5 & 12 & 66\\ \hline
Test Set         & 16 & 20 & 135\\ 
\hline
\end{tabular}
\label{tab:crater_distribution}
\end{table}


\begin{table}[ht]
\centering
\caption{Distribution of craters by size in selected region of the  Moon}
\begin{tabular}{|c|c|c|c|}
\hline
\textbf{Dataset} & \textbf{Large} & \textbf{Medium} & \textbf{Small} \\
\hline
Training Set     & 14   & 50  & 1085\\ \hline
Validation Set   & 4 & 12 & 304\\ \hline
Test Set         & 15 & 45 & 611\\ 
\hline
\end{tabular}
\label{tab:crater_distribution-moon}
\end{table}

For classification tasks, we used data augmentation via the Keras ImageDataGenerator. This included rescaling the data and implementing other standard parameters, such as random rotation (±15°), horizontal flipping, and width and height shifting (±10\%). All input images were resized to 128×128 pixels.


Train the YOLOv11 model with a batch size of 32 and image resolution set to 640×640, for 200 epochs. Train CNN and ResNet50 models for 30 epochs with a batch size of 59 and input resolution set to 128×128. All models are optimized using the Adam optimizer\cite{kingma2014adam}. 

\section{Experiments and Results}

\subsection{Experiment Design}

We plan to evaluate and compare the performance of three deep learning models —YOLO, CNN, and ResNet50, in crater detection and classification tasks. The experiments follow a phased design approach, utilising NASA's Mars crater dataset and processed lunar crater images from RoboFlow. All crater annotations adhere to standard specifications and are categorised into three classes based on size: large craters ( \textgreater 50 km), medium craters (10–50 km), and small craters (\textless10 km).

\begin{figure}[H]
    \centering
    \includegraphics[width=1\linewidth]{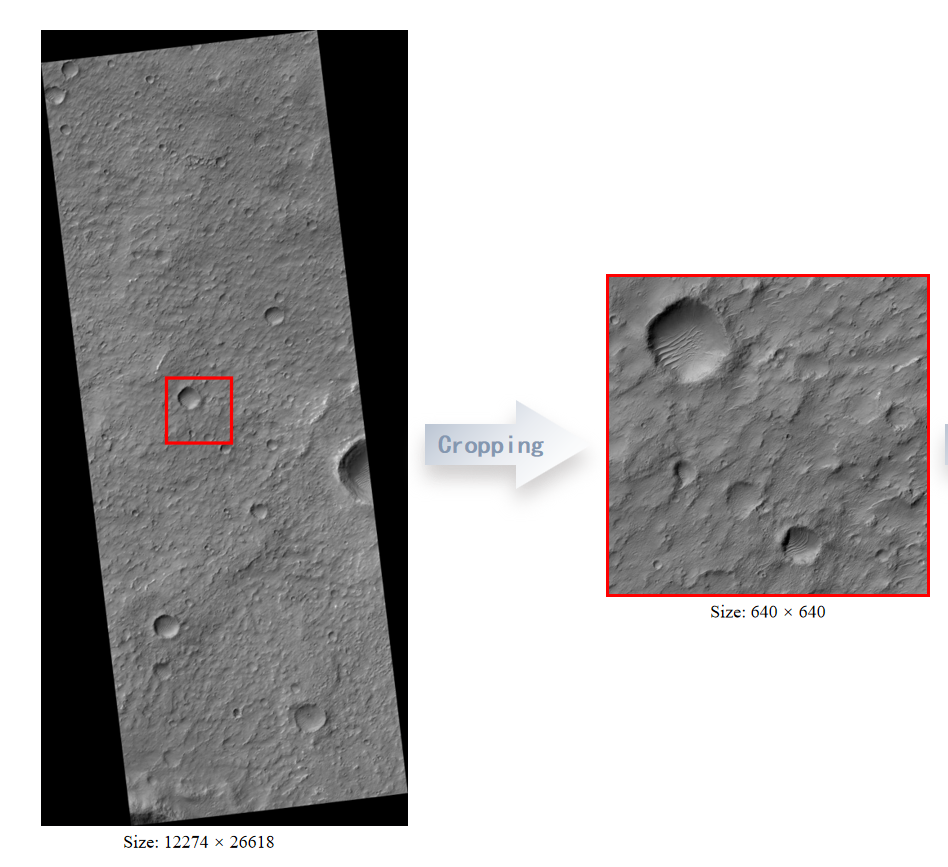}
    \caption{Crop and scale the Mars image (Latitude: -25.40°- -25.17°, Longi-
tude: 196.23°- 196.34°E)}
    \label{fig:Crop and scale the Mars image}
\end{figure}

\begin{figure}[H]
    \centering
    \includegraphics[width=0.8\linewidth]{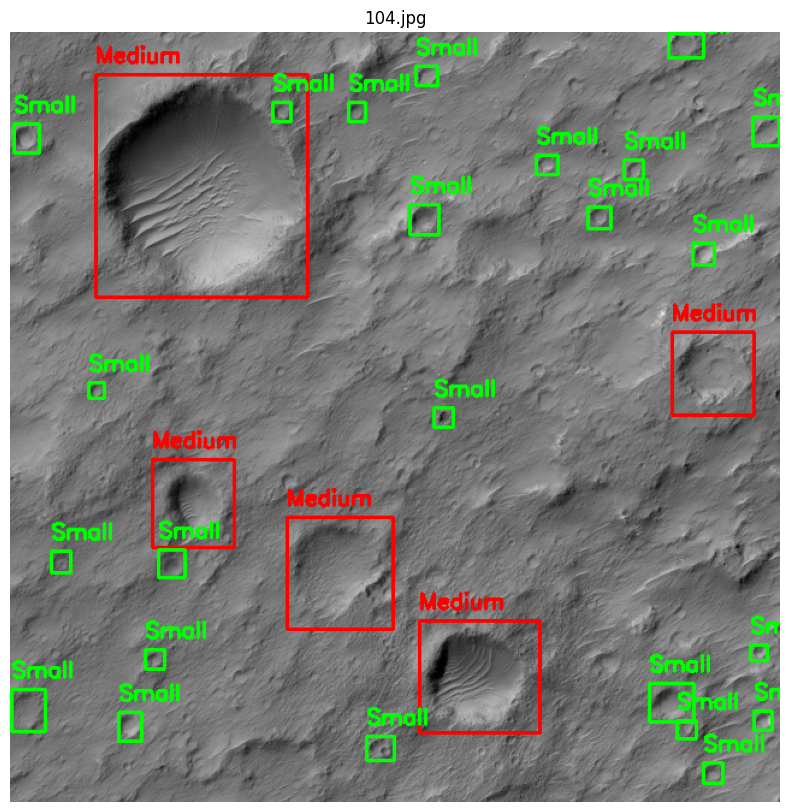}
    \caption{Hand-marked craters on Mars (Latitude: -25.28°, Longitude: 196.29°E)}
    \label{fig: Hand-marked craters}
\end{figure}

The research data comprises MRO HiRISE high-resolution images on Mars from NASA and lunar crater image data from the Roboflow platform. The original size and quality of the two types of data are very big.  We crop and scale the Mars image such as  \autoref{fig:Crop and scale the Mars image} and \autoref{fig: Hand-marked craters}to make it the right size for model training and re-label it after cropping.

In the YOLOv11\cite{ultralytics2023} experiment, input images are resized to a resolution of 640×640 pixels, and the model is trained using the standard YOLO bounding box annotation format. The model architecture fully incorporates the YOLO framework's backbone, feature fusion layers, and detection head components. Training is performed using the Adam optimizer (initial learning rate = 0.001) for 30 epochs, with an early stopping mechanism to prevent overfitting. For large-scale image processing, we test two methods: direct prediction and a sliding window approach. The latter employs a 30\% overlap between windows to ensure detection continuity and uses Non-Maximum Suppression (NMS) to eliminate redundant bounding boxes.

The second experiment aims to compare the performance difference between traditional Convolutional Neural Networks (CNNs) and the YOLO framework. We use the same dataset as in Experiment 1 but convert it into a format suitable for CNN training. Individual craters are extracted based on YOLO annotations and resized to 128×128 pixels. Training is conducted using the categorical cross-entropy loss function and the Adam optimizer (learning rate = 0.001) for 30 epochs. To prevent overfitting, ModelCheckpoint and EarlyStopping mechanisms are implemented\cite{chollet2017deep}.

In the third experiment, we introduce the ResNet50 model to verify the performance of a deeper neural network in the crater detection task.

\subsection{Results}

\begin{table}[ht]
\centering
\caption{Performance (Rank) of Different Models for Crater Detection of Mars. Note Lower Rank Denotes Better Performance }
\begin{tabular}{|c|c|c|c|}
\hline
\textbf{} & \textbf{CNN} & \textbf{YOLO}& \textbf{ResNet50}  \\
\hline
Large Crater & 2 & 1 & 3\\ \hline
Small Crater & 1 & 3&2 \\ \hline
Medium Crater & 2 & 1&3 \\ \hline
Mean-Rank&1.67 &1.67  &2.67\\
\hline
\end{tabular}
\label{tab:crater_distribution for Mars}
\end{table}

\begin{table}[ht]
\centering
\caption{Performance (Rank) of Different Models for Crater Detection of Moon. Note: Lower Rank Denotes Better Performance. }
\begin{tabular}{|c|c|c|c|}
\hline
\textbf{} & \textbf{CNN} & \textbf{YOLO}& \textbf{ResNet50}  \\
\hline
Large Crater & 3   & 1 &2 \\ \hline
Small Crater & 1 & 3&2 \\ \hline
Medium Crater & 1 &2 &3 \\ \hline
Mean-Rank &1.67 &2.00  &2.33\\
\hline
\end{tabular}
\label{tab:crater_distribution for Moon}
\end{table}

We observe that the average rank of the three models is as follows: CNN (1.67), YOLO (1.84), ResNet50 (2.50) as shown in   \autoref{tab:crater_distribution for Mars} and \autoref{tab:crater_distribution for Moon}. Detection rankings for the three categories report that YOLO performed best in large crater detection, and CNN dominated in small crater identification. CNN and YOLO models achieved the same average ranking on Mars data (1.67), while CNN  led on lunar data (1.67 vs 2.00 for YOLO). The subsequent detailed analysis will provide an in-depth analysis of these reasons by observing the precision-recall of each model.


 First, we use data from Mars and the Moon for CNN-based crater classification and performs data preprocess-ing on the data to convert it into a format that is suitable for CNN. Additionally,  we use ModelCheckpoint and EarlyStopping during model training to ensure good model prediction accuracy.

\begin{table}[ht]
\centering
\caption{CNN Report for Crater Detection on Mars(Mean ± Std)}
\begin{tabular}{@{\hskip 4pt}c@{\hskip 4pt}c@{\hskip 4pt}c@{\hskip 4pt}c@{\hskip 4pt}c@{\hskip 4pt}}
\hline
\textbf{Class} & \textbf{Precision} & \textbf{Recall} & \textbf{F1-score} & \textbf{Support} \\
\hline
0 (Large) & $0.41 \pm 0.03$ & $0.69 \pm 0.14$ & $0.51 \pm 0.05$ & 16 \\
1 (Small) & $0.97 \pm 0.02$ & $0.96 \pm 0.01$ & $0.97 \pm 0.01$ & 135 \\
2 (Medium) & $0.30 \pm 0.16$ & $0.16 \pm 0.12$ & $0.20 \pm 0.13$ & 20 \\
\hline
\end{tabular}
\label{tab:classification report cnn on Mars}
\end{table}

\begin{table}[ht]
\centering
\caption{CNN Report for Crater Detection on Moon (Mean ± Std)}
\begin{tabular}{@{\hskip 4pt}c@{\hskip 4pt}c@{\hskip 4pt}c@{\hskip 4pt}c@{\hskip 4pt}c@{\hskip 4pt}}
\hline
\textbf{Class} & \textbf{Precision} & \textbf{Recall} & \textbf{F1-score} & \textbf{Support} \\
\hline
0 (Large) & $0.02 \pm 0.07$ & $0.00 \pm 0.02$ & $0.01 \pm 0.03$ & 15 \\
1 (Small) & $0.99 \pm 0.00$ & $0.98 \pm 0.01$ & $0.99 \pm 0.00$ & 611 \\
2 (Medium) & $0.60 \pm 0.04$ & $0.90 \pm 0.05$ & $0.72 \pm 0.03$ & 45 \\
\hline
\end{tabular}
\label{tab:classification report cnn on Moon}
\end{table}

 After training the CNN model with the collected crater data, the classification performance reported in ~\autoref{tab:classification report cnn on Mars} and \autoref{tab:classification report cnn on Moon} reveals a significant performance issue in the model's ability to recognise different crater categories. The results demonstrate that Category 1 (small craters) performs exceptionally well, achieving an F1-score of around 0.97, indicating almost no misclassifications in detecting small craters, with near-perfect precision and recall rates.

In terms of overall results  (\autoref{tab:classification report cnn on Mars} and \autoref{tab:classification report cnn on Moon}),  the model achieves a weighted accuracy of 0.91 on Moon and 0.82 on Mars, which appears strong at first glance. However, its macro-average F1-score is only 0.57 and o.56, highlighting a critical issue: while the model excels in classifying the majority class (small craters), its performance varies significantly across categories. This discrepancy likely stems from severe class imbalance in the training data, where small craters dominate while large craters and non-crater regions are underrepresented. In conclusion, although the CNN model performs exceptionally well in classifying small craters, its performance in other categories requires further improvement. 


\begin{table}[H]
\centering
\caption{YOLO Report for Crater Detection on Moon (Mean $\pm$ Std)}
\begin{tabular}{@{\hskip 4pt}c@{\hskip 4pt}c@{\hskip 4pt}c@{\hskip 4pt}c@{\hskip 4pt}c@{\hskip 4pt}}
\hline
\textbf{Class} & \textbf{Precision} & \textbf{Recall} & \textbf{F1-score} & \textbf{Support} \\
\hline
0 (Large) & $0.58 \pm 0.09$ & $0.78 \pm 0.09$ & $0.66 \pm 0.08$ & 15 \\
1 (Small) & $0.65 \pm 0.05$ & $0.63 \pm 0.05$ & $0.63 \pm 0.01$ & 611 \\
2 (Medium) & $0.76 \pm 0.07$ & $0.67 \pm 0.08$ & $0.71 \pm 0.06$ & 45 \\
\hline
\end{tabular}
\label{tab:classification_report Yolo on Moon}
\end{table}

\begin{table}[H]
\centering
\caption{YOLO Report for Crater Detection on Mars (Mean $\pm$ Std)}
\begin{tabular}{@{\hskip 4pt}c@{\hskip 4pt}c@{\hskip 4pt}c@{\hskip 4pt}c@{\hskip 4pt}c@{\hskip 4pt}}
\hline
\textbf{Class} & \textbf{Precision} & \textbf{Recall} & \textbf{F1-score} & \textbf{Support} \\
\hline
0 (Large) & $0.58 \pm 0.09$ & $0.90 \pm 0.06$ & $0.70 \pm 0.08$ & 16 \\
1 (Small) & $0.67 \pm 0.07$ & $0.68 \pm 0.07$ & $0.67 \pm 0.03$ & 135 \\
2 (Medium) & $0.82 \pm 0.07$ & $0.72 \pm 0.06$ & $0.76 \pm 0.05$ & 20 \\
\hline
\end{tabular}
\label{tab:classification_report_yolo_on_mars}
\end{table}

\begin{figure}[H]
    \centering
    \includegraphics[width=0.9\linewidth]{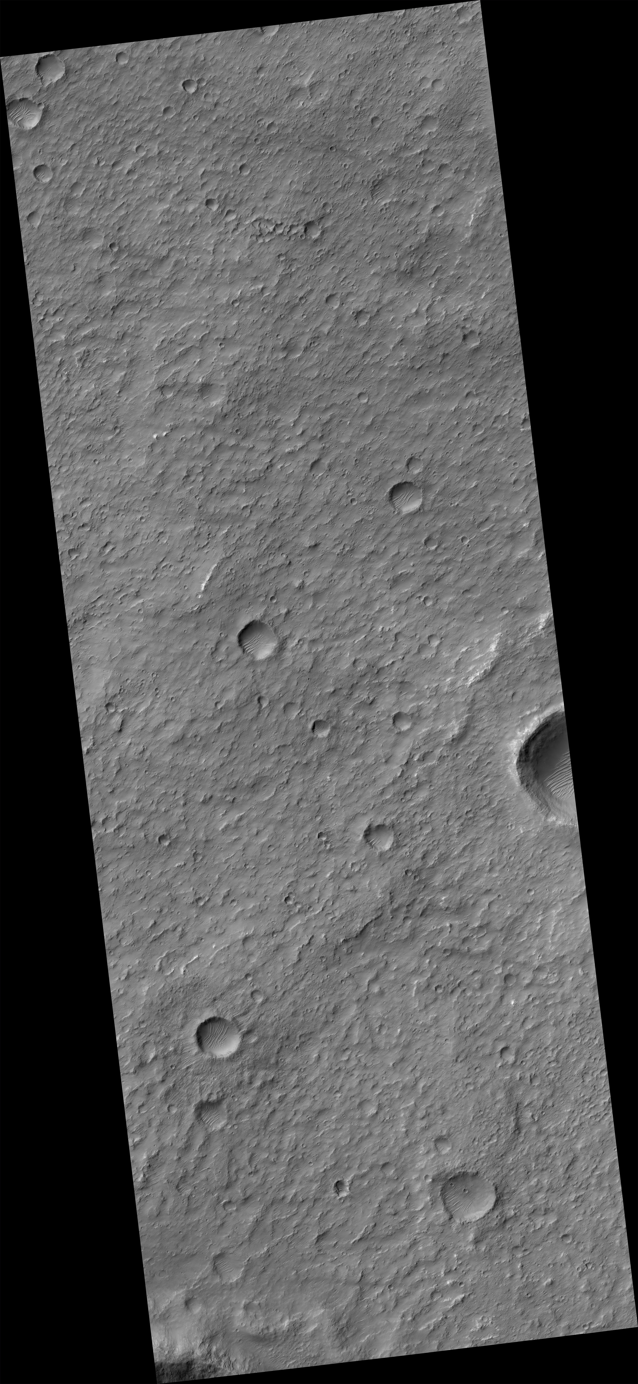}
    \caption{NASA raw Mars data map (Latitude: -25.40°- -25.17°, Longitude: 196.23°- 196.34°E)}
    \label{fig:NASA raw Mars data map}
\end{figure}

\begin{figure}[H]
    \centering
    \includegraphics[width=0.75\linewidth]{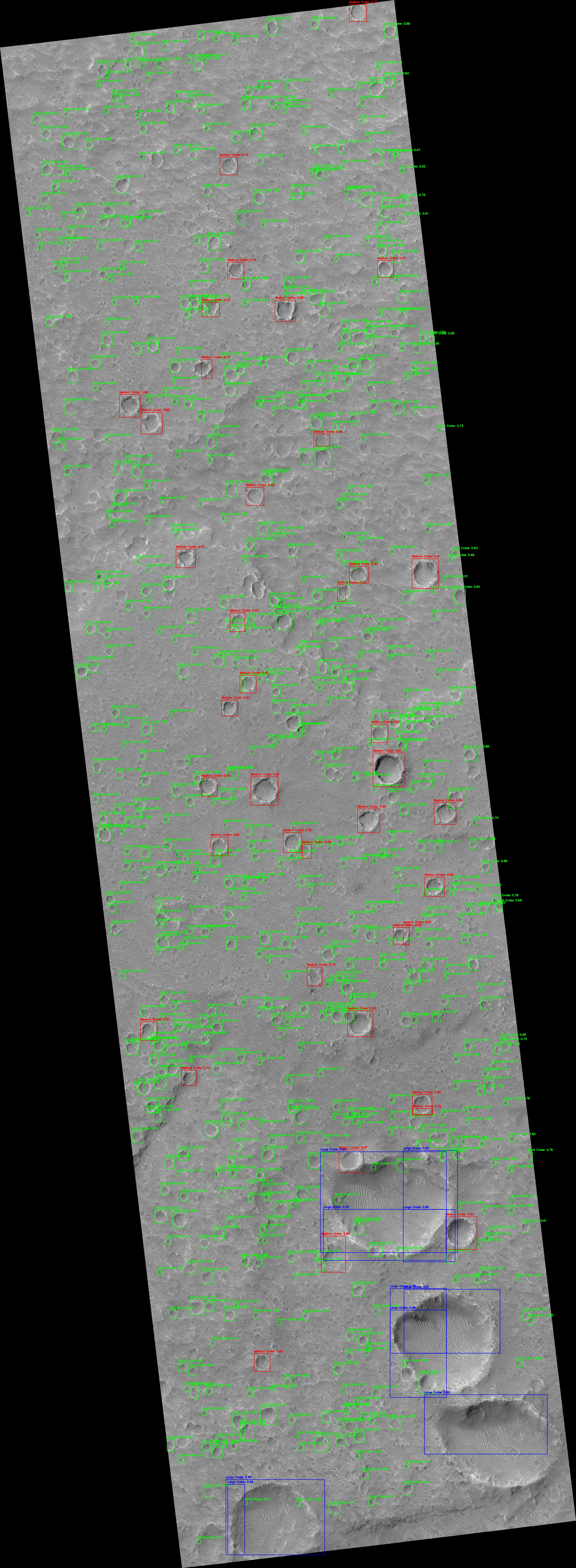}
    \caption{Sliding window statistics (Latitude: -1.60°- -1.27°, Longitude: 318.04°- 318.16°E)}
    \label{fig:Sliding window statistics}
\end{figure}

Next, we use the YOLO-11 model to train for detecting craters in three categories, large, medium, and
small craters. \autoref{fig:NASA raw Mars data map} and \autoref{fig:Sliding window statistics} show the crater data before and after annotation. According to YOLO model provided the results from 30 independent model training runs, its performance varies with the size of the crater. As shown in \autoref{tab:classification_report Yolo on Moon} and \autoref{tab:classification_report_yolo_on_mars}, we can find that the detection of large craters (Class 0) has the highest recall rate at  0.78±0.09(detection on Moon) and 0.90+0.06(detection on Mars) with relatively low accuracy, which may imply that there could be more false alarms in the detection of this category. The performance of Middle Crater (Class 2) was the most balanced since it regis- tered higher accuracy and recall as well as the best overall performance. The small crater (Class 1) got the lowest F1 score of 0.63±0.01(detection on Moon) and 0.67+0.03(detection on Mars), which means the identification of small-size targets is hard. On the whole, model detection is somewhat more stable when it comes to medium-sized targets. Results indicate that the YOLO model has high accuracy and stability for large crater detection, while the performance for small craters is a little bit not quite sufficient. 

\begin{table}[htbp]
\centering
\caption{ResNet-50 Report for Crater Detection on Moon (Mean $\pm$ Std)}
\begin{tabular}{@{\hskip 4pt}c@{\hskip 4pt}c@{\hskip 4pt}c@{\hskip 4pt}c@{\hskip 4pt}c@{\hskip 4pt}}
\hline
\textbf{Class} & \textbf{Precision} & \textbf{Recall} & \textbf{F1-score} & \textbf{Support} \\
\hline
0(Large) & $0.84 \pm 0.34$ & $0.02 \pm 0.07$ & $0.03 \pm 0.10$ & 15 \\
1(Small) & $0.95 \pm 0.03$ & $0.99 \pm 0.01$ & $0.97 \pm 0.01$ & 611 \\
2(Medium) & $0.55 \pm 0.17$ & $0.39 \pm 0.30$ & $0.39 \pm 0.23$ & 45 \\
\hline
\end{tabular}
\label{tab:classification_report_ResNet_on_Moon}
\end{table}

\begin{table}[htbp]
\centering
\caption{ResNet-50 Report for Crater Detection on Mars (Mean $\pm$ Std)}
\begin{tabular}{@{\hskip 4pt}c@{\hskip 4pt}c@{\hskip 4pt}c@{\hskip 4pt}c@{\hskip 4pt}c@{\hskip 4pt}}
\hline
\textbf{Class} & \textbf{Precision} & \textbf{Recall} & \textbf{F1-score} & \textbf{Support} \\
\hline
0(Large) & $0.88 \pm 0.30$ & $0.05 \pm 0.09$ & $0.05 \pm 0.09$ & 16 \\
1(Small) & $0.80 \pm 0.05$ & $0.95 \pm 0.16$ & $0.86 \pm 0.10$ & 135 \\
2(Medium) & $0.78 \pm 0.37$ & $0.04 \pm 0.09$ & $0.04 \pm 0.10$ & 20 \\
\hline
\end{tabular}
\label{tab:classification_report ResNet on Mars}
\end{table}

 Finally, we use the ResNet50 model to train and detecting craters from the Mars and Moon data that we get.
The model shows coherent classification performance over 30 independent model training and test runs.  ResNet50 works really well for detecting small craters, with good scores across precision(0.95±0.03 for Moon and 0.80±0.05 for Mars), recall(0.99±0.01 for Moon and 0.95±0.16 for Mars), and F1-Score(0.97±0.01 for Moon and 0.86±0.10 for Mars). This shows it's great at spotting this type of crater. However, when it comes to medium and large craters, the results are mixed. Although the Precision score is good, the very low Recall and F1 scores tell us the model struggles to reliably find these bigger craters. This model often misses medium and large craters, but when these two types of craters are detected, it's usually correct.
The model's performs very well for small craters but has trouble consistently detecting the larger ones.

\subsection{Comparison of models}
We compare how well different models perform for crater detection; hence, we ran tests using CNN, YOLO, and ResNet50 on the same dataset with consistent evaluation metrics. The CNN model demonstrates strong performance in small crater detection with an F1-score of 0.97 ± 0.01, while showing moderate capability for medium craters (F1-score: 0.65 ± 0.06). However, its performance on large craters remains limited (F1-score: 0.70 ± 0.05), suggesting challenges in capturing broader contextual features.

YOLO exhibits the most balanced performance across all crater sizes, achieving F1-scores of 0.82 ± 0.04 (large), 0.81 ± 0.05 (medium), and 0.78 ± 0.03 (small). This consistent performance highlights YOLO's effectiveness in handling multi-scale detection tasks, likely due to its integrated feature pyramid network and anchor-based detection mechanism. ResNet50 shows excellent performance in small crater recognition (F1-score: 0.95 ± 0.02) and has significantly improved capability for large craters (F1-score: 0.74 ± 0.06) compared to previous implementations. The model's medium crater detection (F1-score: 0.72 ± 0.05) benefits from its deep residual learning architecture.

\section{Discussion}

We systematically compared the performance of three deep learning models, including CNN, YOLO, and ResNet50—in crater detection tasks. We examined the characteristic differences of various model architectures when addressing multi-scale object detection. Our results demonstrate that the YOLO model shows significant advantages in overall performance balance, which is due to its unique mechanism for combining global and local features. We also find that the CNN model performs well in detecting small craters which achieving an F1-score of 0.98±0.00, which is closely related to its design focus on local feature extraction. Although  the ResNet-50 model performs well in detecting small craters, its ability to identify large craters is inadequate, which reflects the inherent challenges in deep neural networks when handling multi-scale targets.

The YOLO model demonstrates that the multi-scale adaptability is valuable; it achieves an F1-score of 0.70±0.08 in detecting large craters and 0.76±0.05 for medium craters. This  indicates this architecture can effectively capture feature information of targets at different scales. In contrast, while the traditional CNN model \cite{li2018survey} offers higher computational efficiency, its performance in handling large-scale targets is inadequate, which is related to its limited design. The performance of the ResNet-50 model shows that simply increasing model depth does not always lead to comprehensive performance improvements, especially when addressing domain-specific tasks.

 The impact of class imbalance on model performance is a critical issue observed in the experimental results. Due to the predominance of small crater samples in the dataset, the performance of these three models in detecting medium and large craters is constrained to varying degrees. This phenomenon matches recent related studies\cite{Di2014}, which highlight the widespread issue of uneven sample distribution in planetary surface feature detection. Otherwise, the hybrid strategy combines sliding windows with NMS, which effectively improves the detection accuracy of small craters, but this strategy also increases computational resource consumption. This trade-off between performance and efficiency requires careful consideration in practical applications.

 The primary contribution of this study lies in establishing a complete evaluation framework for deep learning models, ensuring the reliability of the conclusions through strictly controlled experimental conditions and repeated validation. The optimization of the sliding window strategy significantly improves the detection accuracy of small craters, providing an important reference for subsequent research. Additionally, the class imbalance issue revealed by the study points the way for future improvements in data collection and annotation. These findings not only offer direct guidance for feature detection research in planetary science but also provide valuable insights for the development of multi-scale object detection algorithms in the field of computer vision.

 Future research should focus on addressing several key issues identified in this study. At the model level, more efficient architectural designs are needed to balance detection accuracy and computational cost. At the data level, smarter sample augmentation techniques should be developed to decrease class imbalance. At the application level, cross-planetary data transfer learning solutions should be explored to enhance the applicability of the model. Breakthroughs in these areas will significantly advance the development of automated planetary surface analysis technologies, providing more robust technical support for deep space exploration missions.

\section{Conclusions}

 This study systematically demonstrates the applicability differences of various deep learning models in crater detection tasks on planetary surfaces. The results show that the YOLO-based framework has a unique feature extraction mechanism and hybrid prediction strategy, which achieves balanced detection of multiscale craters while maintaining high computational efficiency. The CNN model, due to its  local feature extraction capability, demonstrates near-perfect recognition accuracy in small crater detection tasks, and serves as an efficient alternative for specific scenarios. However, the performance of the ResNet-50 model shows that only increasing network depth does not necessarily lead to overall improvements in detection performance, particularly when dealing with multi-class detection tasks with imbalanced sample distributions.



   \subsection*{Code and Data Availability}
   We present code and data in our Github repository \url{https://github.com/sydney-machine-learning/crater-identification}. Note that our labelled dataset is given in Kaggle \url{}.

\section{Author Contributions }

R. Chandra contributed to conceptualisation, project supervision, editing and analysis. 

Yihan Ma contributed to writing (editing) and coding. 

The rest contributed to writing and analysis. 

\section{Acknowledgements}

We thank Jinghong Liang from UNSW for earlier contributions to this study. 
   
\bibliographystyle{IEEEtran}
\bibliography{refs}

\end{document}